\documentclass[conference]{IEEEtran}
\IEEEoverridecommandlockouts
\usepackage{cite}
\usepackage{amsmath,amssymb,amsfonts}
\usepackage{algorithmic}
\usepackage{graphicx}
\usepackage{textcomp}
\usepackage{xcolor}
\usepackage{mathtools}
\def\BibTeX{{\rm B\kern-.05em{\sc i\kern-.025em b}\kern-.08em
    T\kern-.1667em\lower.7ex\hbox{E}\kern-.125emX}}
\begin{document}

\title{D-SRGAN: DEM Super-Resolution with Generative Adversarial Networks\\
\thanks{}
}

\author{\IEEEauthorblockN{\textsuperscript{} Bekir Z Demiray}
\IEEEauthorblockA{\textit{Department of Computer Science} \\
\textit{University of Iowa}\\
Iowa City, USA \\
bekirzahit-demiray@uiowa.edu}
\and
\IEEEauthorblockN{\textsuperscript{} Muhammed Sit}
\IEEEauthorblockA{\textit{Graduate Program in Informatics} \\
\textit{University of Iowa}\\
Iowa City, USA \\
muhammed-sit@uiowa.edu}
\and
\IEEEauthorblockN{\textsuperscript{} Ibrahim Demir}
\IEEEauthorblockA{\textit{Department of Civil \& Environmental Engineering} \\
\textit{University of Iowa}\\
Iowa City, USA \\
ibrahim-demir@uiowa.edu}
}

\maketitle

\begin{abstract}
LIDAR (light detection and ranging) is an optical remote-sensing technique that measures the distance between sensor and object, and the reflected energy from the object. Over the years, LIDAR data has been used as the primary source of Digital Elevation Models (DEMs). DEMs have been used in a variety of applications like road extraction, hydrological modeling, flood mapping, and surface analysis. A number of studies in flooding suggest the usage of high- resolution DEMs as inputs in the applications improve the overall reliability and accuracy. Despite the importance of high-resolution DEM, many areas in the United States and the world do not have access to high-resolution DEM due to technological limitations or the cost of the data collection. With recent development in Graphical Processing Units (GPU) and novel algorithms, deep learning techniques have become attractive to researchers for their performance in learning features from high-resolution datasets. Numerous new methods have been proposed such as Generative Adversarial Networks (GANs) to create intelligent models that correct and augment large-scale datasets. In this paper, a GAN based model is developed and evaluated, inspired by single image super-resolution methods, to increase the spatial resolution of a given DEM dataset up to 4 times without additional information related to data.
\end{abstract}


\section{Introduction}
Digital Elevation Models (DEMs) are visualization of terrain’s surface powered by elevation data. Over the years, DEMs have served as input in many research studies including stream network extraction \cite{b33}, flood risk and hazard mapping \cite{b34}, extracting urban features \cite{b35}, surface texture analysis \cite{b36}, and generation of drainage networks \cite{b37}. Due to a variety of application areas, the generation of DEMs has been studied in many fields with different techniques such as conventional topographic surveys, kinematic GPS surveys, radar techniques, analogue and laser surveys \cite{b38}. Another approach for DEM generation is LiDAR (light detection and ranging). LiDAR is an optical remote-sensing technique that measures the distance between sensor and object, and reflected energy from the object. LiDAR data has been used as the primary source of high-resolution and accurate DEMs. Despite widely usage of LiDAR data, DEMs still contain issues and systematic errors \cite{b39, b40, b41}. The process of generating DEM consists of numerous steps including data collection, data reduction, interpolation, and etc. Each of these steps contains some level of uncertainties and accumulation of these uncertainties has a plausible effect in revealing the desired quality level of DEM
\cite{b42, b43}. The resolution and accuracy of DEM have a significant effect on the outcome of these analyses. The resolution of DEM is very crucial for many applications. It is shown that the resolution and information content of DEM has a massive impact on the computed topographic indices \cite{b44}. Chaubey et al. \cite{b45} examined the effect of DEM resolution on the predictions from the SWAT (Soil and Water Assessment Tool) model and they found a clear link between DEM resolution and accuracy of predicted stream network and sub-basin classification in the SWAT model. Similarly, several studies demonstrated that using high-resolution DEMs as inputs, construct more accurate flood maps compared to low-resolution DEMs \cite{b85, b86}.
\\
\hspace*{5mm}Water resource management and hydrological modeling using physically-based or data-driven (i.e. artificial neural networks) approaches \cite{b69, b70} need high-resolution DEM for accurate hydrological predictions \cite{b46}. Web-based systems for efficient disaster management, recovery and response, interactive flood visualizations \cite{b71, b72}, information platforms \cite{b73, b74, b75}, and intelligent systems \cite{b76, b77} relies on high quality DEM. Besides advanced hydrological modeling, monitoring and geographic analysis such as watershed delineation \cite{b78, b82} and stage height measurements \cite{b79} benefit from DEMs. In some cases, it is not feasible to use high-resolution DEM due to the limitation of computing systems or model run time. Even in such cases, resampling high-resolution DEM to lower one gives a better result than the original coarse resolution DEM. Despite the importance of high-resolution DEM, many areas in the United States and the world do not have access to high-resolution DEMs due to technological limitations or the cost of the data collection process \cite{b47}. As an alternative, enhancing the resolution (super-resolution) of the existing datasets can be seen as the optimal approach to fill the gap. Super-resolution is a widely studied topic in computer vision in which aims to generate high-resolution images with the help of one or multiple low-resolution images. DEM, denoted by a matrix, is highly similar to images in terms of denotation. DEM could be considered as an image in super-resolution application, since its planer coordinates and height values can be seen as the pixel position and corresponding color values, respectively \cite{b48}.
\\
\hspace*{5mm}With recent developments in Graphical Processing Units (GPU) and novel algorithms, deep learning techniques have become attractive to researchers for their performance in learning features. For example, neural networks are used in the field of hydrology to develop flood forecasting models \cite{b69, b70}, realistic river imagery generation \cite{b80} and disaster related tweet classification \cite{b81}. Convolutional Neural Networks (CNNs), a deep neural network algorithm, based single-image super-resolution (SRCNN) demonstrated the effectiveness of CNNs in image enhancement \cite{b14}. Taking advantage of the similarity between DEM datasets and images, D-SRCNN is developed in this study to increase the resolution of DEMs with similar approaches in SRCNN \cite{b49}. Alongside the success of CNN, new deep neural network algorithms like Generative Adversarial Networks (GANs) have been started to gain attention in super-resolution literature. SRGAN, one of the early successful examples of GANs in super-resolution, achieved to increase the resolution of an image up to four times upscaling factor with high performance \cite{b23}.
\\
\hspace*{5mm}In this paper, the power of GANs is explored to develop a deep neural network model, D-SRGAN, that aims to convert provided low-resolution DEMs into high-resolution ones without additional information. The performance of D-SRGAN is compared with traditional interpolation methods such as bicubic and bilinear in order to understand the effectiveness of the approach. The paper is organized as follows: Section 2 reviews the relevant research in image super-resolution and DEM super-resolution. Section 3 gives details about our network design and the general concept. Also, experimental data is provided in Section 3. Section 4 covers the detailed results of the proposed method and related discussion. The paper finalized with conclusions and possible future research paths. 

\section{Related Work}
Super-resolution is a process of producing a high-resolution image from one or more low-resolution images and it is one of the active fields in computer science. Super-resolution can be classified into two groups: multi-frame super-resolution and single image super-resolution (SISR) \cite{b1, b2}. Multi frame super-resolution combines information from different low-resolution images to produce a higher resolution image by employing various techniques such as iterative back projection or probabilistic approaches \cite{b56, b57, b58, b59, b60}. Since our concept focuses on single image super-resolution, we will not provide further information about multi-frame super-resolution. Over the years, various approaches have been proposed on SISR. Interpolation-based methods such as linear, bicubic or Lanczos have been applied with the power of predefined mathematical formulation without the training phase. Despite the performance of these methods, they underperform at high-frequency regions due to the tendency of smoothness \cite{b2, b3, b4}. Reconstruction-based methods take advantage of prior knowledge to generate high-resolution images. Various approaches have been used in reconstruction-based methods such as steering kernel regression (SKR) \cite{b61} or non-local means (NLMs) \cite{b62}. Alongside the success of preserving edges and suppressing artifacts, reconstruction-based methods are not successful in producing super-resolution images at large magnification factors. \cite{b5, b6, b7, b8} Learning or example-based methods aim to gather insight information from paired low and high-resolution images to understand missing details in low-resolution images. Numerous approaches have been proposed as learning or example-based methods such as neighbor embedding \cite{b10}, sparse coding \cite{b11} and regression methods \cite{b12, b13}. One of the crucial elements for these methods is the training set. Quality of the training set can lead to capture redundant or erroneous features and reduce the effectiveness of the methods dramatically \cite{b1, b2, b6, b9}.
\\
\hspace*{5mm}Despite the fact that the root of Convolutional Neural Networks goes back \cite{b83, b84}, CNNs are starting to reach its true potential with the help of recent developments on modern GPUs (Graphic Processing Units). Several novel approaches have been used in different tasks such as image classification \cite{b63}, face recognition \cite{b64}, super-resolution \cite{b15}, object detection [66]. In the literature regarding SISR, Dong et al. \cite{b14} have proposed a method namely Super-Resolution Convolutional Neural Network (SRCNN) to learn a mapping end to end between the low and high-resolution images. The method starts with bicubic interpolation of low-resolution image followed by overlapping patch extraction and representation as high-dimensional vector, then non-linearly maps the high-dimensional vector to another high-dimensional vector, finally it reconstructs the high-resolution image from these vectors. Fast Super-Resolution Convolutional Neural Network (FSRCNN) has been developed by Dong et al. \cite{b15} to increase speed of current SRCNN. In the FSRCNN, deconvolutional layer has been chosen over bicubic interpolation and single mapping layer replaced with four mapping layers and an expanding layer. Kim et al. \cite{b16} constructed a network powered by 20 convolutional layers with a high learning rate and the result of that network is considerably better in comparison to the methods at that time. Deep-Recursive Convolutional Network (DRCN) \cite{b17} is powered by deep recursive layers. Accuracy of the model can be increased with more iteration and it does not require introducing new parameters for additional convolutional layers. DRCN proposed two methods to enhance learning procedure, namely supervision of recursions and skip-connection. Shi et al. \cite{b18} introduced the first convolutional neural network (CNN) capable of real-time SR of 1080p videos on a single K2 GPU. The network consists of L layers. First L-1 layers, feature maps are extracted at low-resolution (LR) space. The final layer, sub-pixel convolutional layer, upscales LR feature maps to high-resolution (HR) output. The study demonstrated that working on the LR space dramatically reduces computational and memory-wise complexity.
\\
\hspace*{5mm}Alongside CNNs, new promising approaches have been explored in super-resolution applications such as Generative Adversarial Networks (GAN). GANs can be considered as a framework that consists of two neural networks designed to defeat each other in a zero-sum game \cite{b19}. After it was proposed, numerous variations of GANs have been tested for various tasks such as image to image translation \cite{b20}, image editing \cite{b21}, biological image synthesis \cite{b22} and random image generation \cite{b80}. SRGAN (Super-Resolution Generative Adversarial Network) is one of the first implementation of GAN designed to achieve SISR. The generator of SRGAN starts with taking the power of deep residual blocks with skip-connections. At the end of the network, the resolution of the image is increased with two sub-pixel convolutional layers. SRGAN uses perceptual loss that consists of adversarial and content losses. Instead of using pixel-wise MSE (Mean Square Error), the content loss is calculated from feature maps of VGG  network (pretrained network by Oxford’s Visual Geometry Group) \cite{b23}. The design of ProGanSR has been influenced by curriculum learning, which proposes the direction of learning should be from small upscaling factors to large upscaling factors. ProGanSR uses the asymmetric pyramids structure to obtain efficiency. Each pyramid consists of Dense Compression Units followed by sub-pixel convolution layers to increase the resolution of input by two times \cite{b24}. Mahapatra et al. proposed local saliency maps, which define the importance of each pixel, to use in the GAN loss function over classical MSE \cite{b25}. Alongside the mentioned papers, numerous papers used GAN to increase the resolution of given image input \cite{b26, b27, b28, b29}.
\\
\hspace*{5mm}The literature on single DEM super-resolution, Xu et. al (2015) proposed a non-local algorithm that searches similar patches over the training set with a predefined equation, then increases the resolution of target DEM with weights calculated through the searching phase \cite{b48}. D-SRCNN is a CNN based method that aims to increase the resolution of given DEM with similar architecture in SRCNN and it performs better than the non-local based method \cite{b49}. Alongside D-SRCNN, Xu et. al \cite{b52} also proposed a CNN based model that is broadly derived from EDSR (Enhanced Deep Super-Resolution Network) \cite{b53}. The network is pre-trained with natural images in order to obtain high-resolution gradient maps which will be fine-tuned with high-resolution DEMs in the next process.In addition to deep learning methods, traditional interpolation approaches such as bicubic, kriging, inverse distance weighting can be used for single DEM super-resolution. Nevertheless, these statistical models tend to produce more smooth terrains [54]. Also, it is possible to use additional data in order to increase the resolution of DEMs. Argudo et al. \cite{b54} proposes a fully convolutional neural network that accepts the low-resolution DEM and its high-resolution orthophoto in order to produce the high-resolution of DEM. Yu et al. \cite{b55} introduces a regularized framework that enables the combine multiple data for corresponding DEM in order to reconstruct a higher resolution DEM. Despite the importance of DEM, the research on single DEM super-resolution is still limited. Recent methods in image super-resolution can be applied to DEM image enhancements with the help of the similarity between DEM and image data.
\section{Methods}
\begin{figure}[htbp]
\centerline{\includegraphics[scale=0.50]{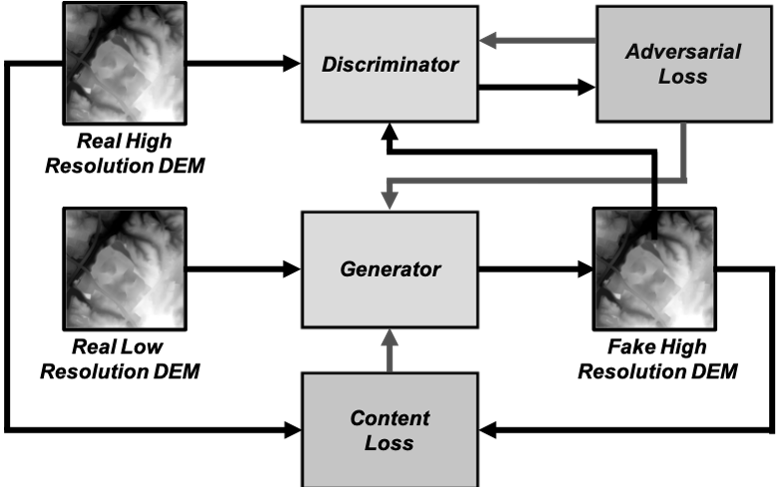}}
\caption{General structure of the D-SRGAN training process.}
\label{fig1}
\end{figure}

Generative Adversarial Networks (GANs) have been used by many researchers from various fields since they were first proposed by Ian Goodfellow et al. (2014) \cite{b19}. GANs consist of two adversarial components, namely Generator and Discriminator, which aim to compete in a minimax game. Generator aims to capture data distribution and produce realistic samples to convince discriminator as fabricated ones are real. On the other hand, discriminator intents to determine the source of incoming samples. The cost of each network is directly related to the success of the opposing component, and the general process can be expressed by the following formulation \eqref{eq1} where discriminator and generator try to beat one another with value function V(G, D).
\begin{equation}
\begin{multlined}
min_{G}min_{D}V(D, G) = \\ \hspace*{6mm} E_{x \sim p_{data^{(x)}}}[logD(x)] + E_{z \sim p_{z^{(z)}}}[log(1 - D(G(z))]
\label{eq1}
\end{multlined}
\end{equation}
\hspace*{5mm}In our study, the goal is generating a high-resolution DEM from a low-resolution DEM. The low-resolution DEMs are accepted by the generator in order to produce high-resolution DEMs. Discriminator of the network takes fabricated or real high-resolution DEM as input and guesses the origin of input. In addition to the adversarial losses during the training, content loss of fabricated high-resolution DEMs is used in the manipulation of the generative network’s weights. The general structure of the GAN training process is represented in Fig. \ref{fig1}.

\subsection{Network Architecture}
\begin{figure}[htbp]
\centerline{\includegraphics[scale=0.50]{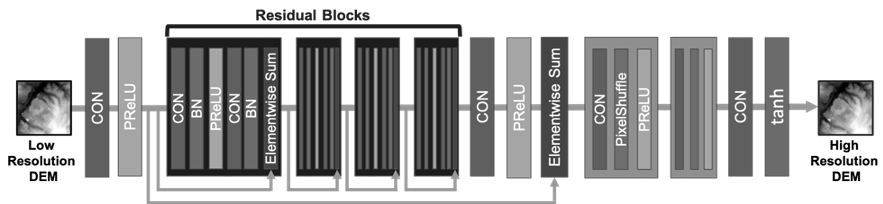}}
\caption{Architecture of generator.}
\label{fig2}
\end{figure}

\begin{figure}[htbp]
\centerline{\includegraphics[scale=0.50]{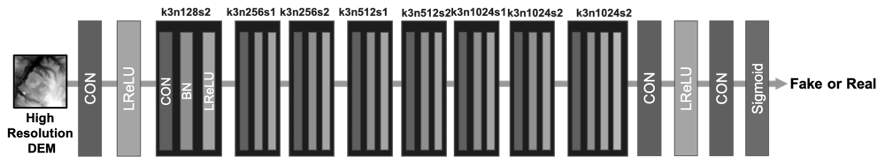}}
\caption{Architecture of discriminator.}
\label{fig3}
\end{figure}

Our network design consists of two opposing components (i.e. generator, discriminator). The architecture of components is based on the SRGAN model \cite{b23}. The generator of our network takes the low-resolution DEM as input and passes it to a convolution layer with 128 feature maps followed by ParametricRelu \cite{b30} as an activation function, then passes it to residual blocks. The generator has eight residual blocks with duplicated design. Each residual block is created with two convolutional layers with 3*3 kernel and 128 feature maps followed by batch normalization layers \cite{b31} and ParametricRelu. Inside each residual block, there is a connection between incoming data from the predecessor component and the last phase of the current residual block which aims to gather low-level features in order to improve the performance of the generator \cite{b32, b17}. A similar link between input data and the output of the final residual block is also established. The next components of the generator are two upsampling blocks which are used to increase the resolution. Upsampling blocks are obtained with a convolutional layer with 512 feature maps followed by sub-pixel convolutional layers [18] and ParametricRelu. In the end, the output of the upsampling blocks is passed to a convolutional layer prior to the tanh function. The visual representation of the generator is provided in Fig. \ref{fig2}.
\\
\hspace*{5mm}The discriminator of the network has nine convolutional layers with 3x3 filter kernels and increment in feature maps from 128 to 1024 by a factor of two. Each convolutional layer is followed by Leaky RELU as an activation function with alpha is equal to 0.2. Strided convolutional layers are used to reduce the resolution of DEM while the number of features is doubling. In the last convolutional layer, there is an additional adaptive average pooling layer prior to dense layers. The outcome of the discriminator is produced with sigmoid function after the dense layers. The visual representation of the discriminator is provided in Fig. \ref{fig3}.

\subsection{Loss Functions}
Under this section, we will review the loss functions applied in the neural network. In the training phase, the adversarial loss is used by both discriminator and generator. In addition to adversarial loss, the generator is affected by content loss in order to converge fast and produce more accurate data points.
\subsubsection{Adversarial Loss}
Adversarial loss is an essential part of the GAN structure. In our design, it is the only element that is used by discriminator of the network as a loss function. The adversarial component is helping to enhance the discriminator while distinguishing the source of data as expected. In the training phase, adversarial loss of discriminator \eqref{eq2} calculated with the following formulation:
\begin{equation}
\begin{multlined}
I_{Dis}^{SR} = \frac{1}{2m}\sum_{i=1}^{m}\mid\mid 1 - D(y_{i})\mid + D(G(x_{i})\mid\label{eq2}
\end{multlined}
\end{equation}
\\
\hspace*{5mm}Adversarial loss is also used for the generator to create more realistic examples with aiming to fool the discriminator. Formulation of adversarial loss for generator \eqref{eq3} calculated as follows:
\begin{equation}
\begin{multlined}
I_{A_Gen}^{SR} = \frac{1}{m}\sum_{i=1}^{m}\mid 1 - D(G(x_{i}))\mid \label{eq3}
\end{multlined}
\end{equation}

\subsubsection{Content Loss}
Alongside adversarial loss, it is a common procedure to use a loss function to determine the difference between ground truth and fabricated data to capture the goodness of produced data and mean square error (MSE) is the widely used optimization value in various work \cite{b48, b49, b54}. It sounds reasonable to use MSE in order to understand the result of the ongoing process, since DEM contains the numerical value of earth surface elevations and MSE is used as a metric to understand the goodness of methods in the field.
\begin{equation}
\begin{multlined}
I_{x}^{SR} = \frac{1}{m}\sum_{i=1}^{m} (x_{i} - y_{i})^2 \label{eq4}
\end{multlined}
\end{equation}

\hspace*{5mm}Since the generator is affected by multiple loss functions, its loss function \eqref{eq5} combination of content loss \eqref{eq4} and adversarial loss \eqref{eq3} and it is as follows:
\begin{equation}
\begin{multlined}
I_{Gen}^{SR} = I_{x}^{SR} + \alpha I_{A_Gen}^{SR} \label{eq5}
\end{multlined}
\end{equation}

\subsection{Data Processing}
\begin{figure}[htbp]
\centerline{\includegraphics[scale=0.50]{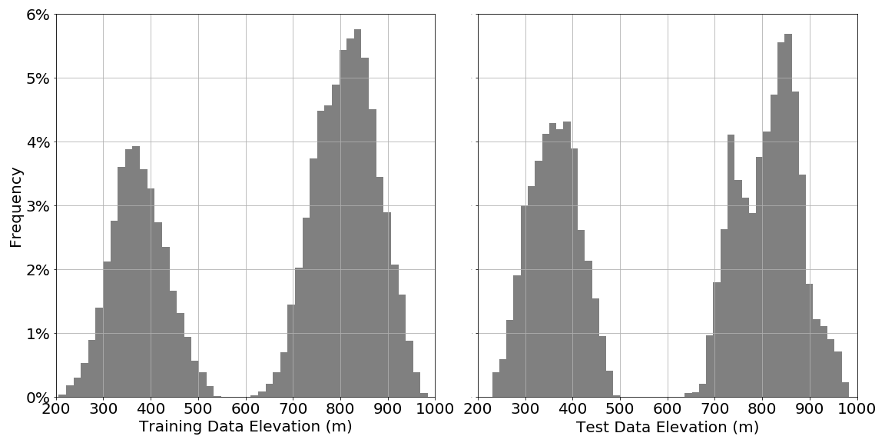}}
\caption{Distribution of elevation data.}
\label{fig4}
\end{figure}

The dataset used in the experiment is collected from North Carolina Floodplain Mapping Program. The dataset covers a total area of 732 km2 from Wake and Guilford counties. As a training set, a total area of 590 km2 is used. The rest of the dataset, area of 142 km2, is accepted as a test set. Each of the used DEMs was collected at a spacing of approximately 2 points per square meter. In the experiment, 3 feet and 50 feet DEMs are used as high-resolution and low-resolution examples, respectively. The NC Program delivered each tile of high-resolution DEMs as 1600x1600 data points and the low-resolution DEMs as 100x100 data points. In the preprocessing phase, HR DEMs are split to 400x400 data points and LW DEMs are split to 25x25 data points. In addition to the fragmentation process, DEMs with missing values are discarded from the dataset prior to the experiment. The average, minimum and maximum elevation values in both datasets are provided in Table \ref{tab1}. The distribution of elevation values is also provided for both datasets in Fig. \ref{fig4}. The network in our study is implemented with Pytorch framework. 

\begin{table}[htbp]
\caption{Statistical summary of elevation datasets for training and testing (m)}
\begin{center}
\begin{tabular}{|c|c|c|c|}
\hline
\textbf{} & \textbf{Avg. Elevation} & \textbf{Min. Elevation} & \textbf{Max. Elevation} \\ \hline
Training  & 653.1                      & 205.7                      & 984.9                      \\ \hline
Test      & 621.7                      & 230.0                      & 982.7                      \\ \hline

\end{tabular}
\label{tab1}
\end{center}
\end{table}

\section{Results and Discussions}
\begin{figure}[htbp]
\centerline{\includegraphics[scale=0.50]{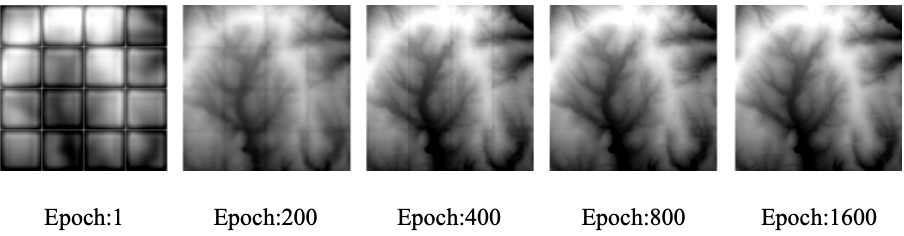}}
\caption{Example outputs of generator during different training epochs.}
\label{fig5}
\end{figure}

The goal of our network is increasing the resolution of given DEM with 4x upscaling factor. The generator is designed to take low-resolution DEMs (50 feet) with 25x25 cells as input and returns high-resolution DEMs (3 feet) with 400x400 cells as output. The discriminator of the network is accepting DEMs with 400x400 cells as input and guess the source of it whether generated by the generator or not. At the beginning of the training procedure, the Adam algorithm \cite{b50} is used as optimizer with learning rates 0.0001 and 0.0002 for discriminator and generator, respectively. The rest of the parameters are accepted with their default values in Pytorch implementation. Also, the weight of adversarial loss in the generator is set to 0.0001. Until the 400th epoch, the discriminator is trained by both fabricated and real HR DEMs. After that point, the training procedure of discriminator is frozen as a favor to the generator, since its performance reaches almost the perfect level. After the 800th epoch, the learning rate of the generator is decreased to 0.0001. In the 1200th epoch, lock on the discriminator is disintegrated but the weight of adversarial loss in the generator is set to 0.00001 due to minimal differences in content loss. This setup remained unchanged until the 2000th epoch where the training of discriminator was frozen again. Fig. \ref{fig5} showcases the visualization of fabricated HR DEM for different epochs with same input DEM.

\begin{figure}[htbp]
\centerline{\includegraphics[scale=0.50]{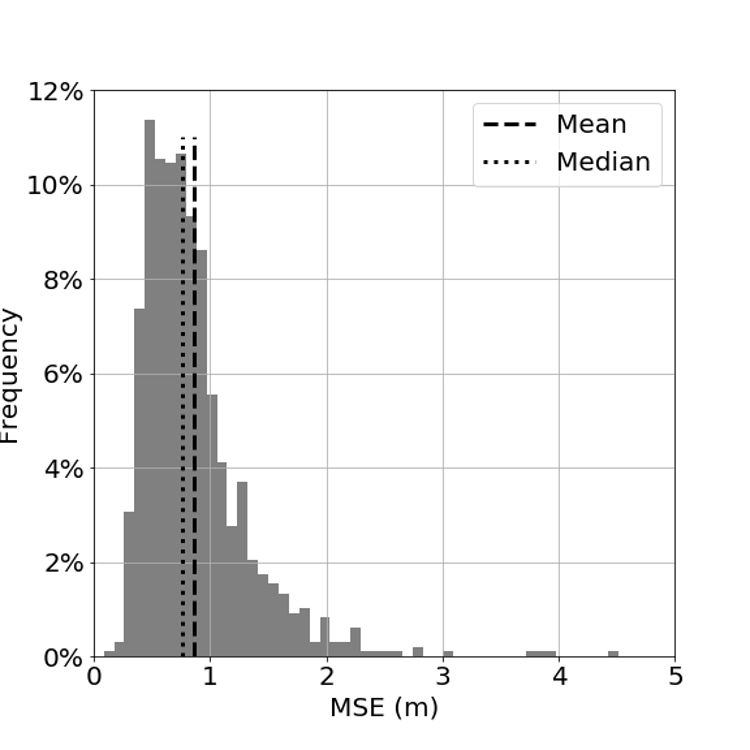}}
\caption{Error distribution of testing dataset.}
\label{fig6}
\end{figure}

Based on the latest network, our GAN based approach provides promising results. Since DEM contains the height value of the corresponding area, it is reasonable to use a metric that reflects quantitative measurements in order to understand the performance of the method. Also, it is common practice to use MSE to understand the effectiveness of proposed methods \cite{b48, b49, b54}. Table \ref{tab2} shows the comparison of different methods on both training and test datasets. According to the results, D-SRGAN outperforms all other tested methods on both training and test datasets. It improves the quality of the images by nearly \%10 in comparison to the best alternative method. The error distribution of results is provided in Fig. \ref{fig6}. The mean, median and standard deviation of error distribution in the testing dataset for D-SRGAN are 0.86, 0.76 and 0.46 meters, respectively. The distribution of errors shows that most of the time D-SRGAN provides promising results with a limited number of outliers.
\begin{table}[htbp]
\caption{Performance comparison of results from D-SRGAN and other methods (m)}
\begin{center}
\begin{tabular}{|c|c|c|c|}
\hline
\textbf{MSE (m)} & \textbf{Training} & \textbf{Test} \\ \hline
D-SRGAN          & 0.865             & 0.861         \\ \hline
Bicubic          & 0.968             & 0.946         \\ \hline
Bilinear         & 1.141             & 1.124         \\ \hline
\end{tabular}
\label{tab2}
\end{center}
\end{table}

\begin{figure}[htbp]
\centerline{\includegraphics[scale=0.50]{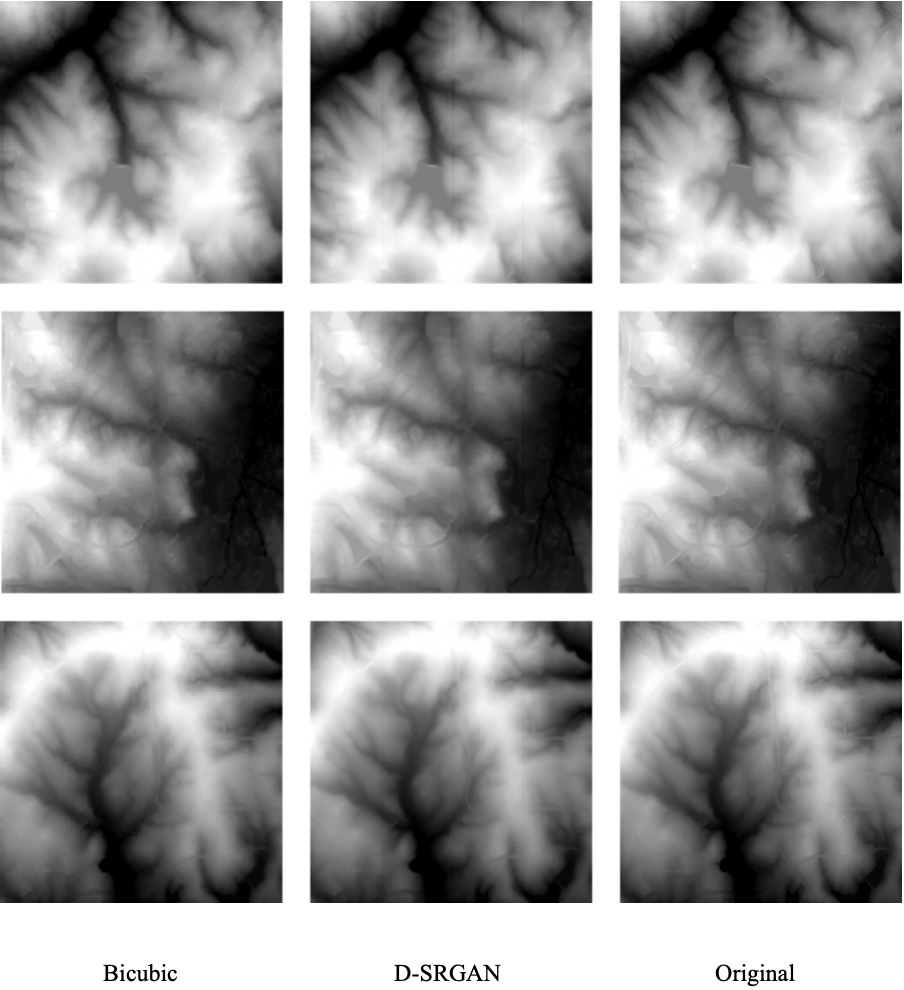}}
\caption{Example SR results from D-SRGAN and bicubic.}
\label{fig7}
\end{figure}

Fig. \ref{fig7} visualizes the example DEMs from the testing dataset which are generated with D-SRGAN, bicubic and bilinear interpolation in order to show the strength and weaknesses of D-SRGAN. D-SRGAN is capable of regeneration of DEMs with 4x higher resolution under the promising deviation. According to Fig. \ref{fig6}, \%81 of all values fall within plus or minus one standard deviation from the mean. As seen from the figure that D-SRGAN is struggling to capture finer details of DEMs. In image super-resolution, MSE based solutions have a tendency to miss high-frequency content and produce more smooth results in which similar effects can be found in D-SRGAN \cite{b23}. However, D-SRGAN still outperforms other methods in similar conditions.

\begin{figure}[htbp]
\centerline{\includegraphics[scale=0.50]{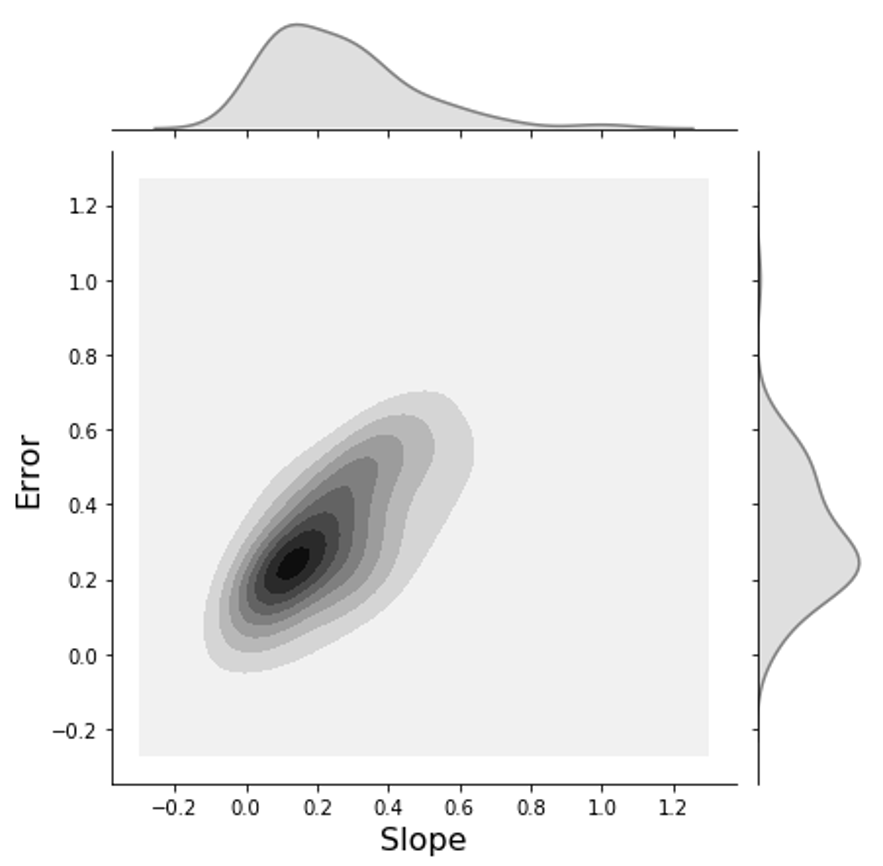}}
\caption{The slope analysis on test set.}
\label{fig8}
\end{figure}

In addition to previous results, slope analysis provides valuable insights regarding the performance of methods in different terrains. The slope is a common parameter that is used in various applications in environmental sciences via DEMs \cite{b66, b67, b68}. For each elevation value of a DEM, the slope is calculated with the average maximum technique proposed by Burrough et al. \cite{b51} based on a 3x3 cells around the value cell. As the slope value changes from lower to greater, the terrain goes from flatter to steeper. Fig. \ref{fig8} shows the slope imagery of the test set as well as the error distribution over slope values which are normalized into [0, 1]. As seen above, D-SRGAN performs better results on flatter terrain than steeper terrain.

\section{Conclusions}
In this study, a generative adversarial network, D-SRGAN, is proposed. D-SRGAN aims to convert low-resolution DEMs into high-resolution ones without needing additional information. The experiment outcomes show that D-SRGAN produces promising results while constructing 3 feet high-resolution DEMs from 50 feet low-resolution DEMs. Despite the overall success of D-SRGAN, D-SRGAN could not perform evenly over the terrains. It produces more realistic examples in flatter terrains than stepper terrains. We believe that this problem can be overcome by using different metrics in the training phase of the generator.

\end{document}